\documentclass[11pt,a4paper]{article}

\usepackage[margin=1in]{geometry}
\usepackage{amsmath,amssymb}
\usepackage{graphicx}
\usepackage{booktabs}
\usepackage{multirow}
\usepackage[colorlinks=true,citecolor=blue,linkcolor=blue,urlcolor=blue]{hyperref}
\usepackage{xcolor}
\usepackage{subcaption}
\usepackage{float}
\usepackage{natbib}
\usepackage{authblk}
\usepackage{setspace}
\usepackage{microtype}

\title{\textbf{Scaling Laws in the Tiny Regime: How Small Models Change Their Mistakes}}

\author[1,3]{Mohammed Alnemari\thanks{Corresponding author: mnemari@ut.edu.sa}}
\author[4]{Rizwan Qureshi}
\author[2]{Nader Begrazadah}

\affil[1]{Faculty of Computer Science and Information Technology, University of Tabuk, Tabuk, Saudi Arabia}
\affil[2]{University of California, Irvine, Irvine, CA, USA}
\affil[3]{AIST Research Center, University of Tabuk, Tabuk, Saudi Arabia}
\affil[4]{Department of Computer Science, Salim Habib University, Karachi, Pakistan}

\date{\today}

\begin{document}

\maketitle

\begin{abstract}
Neural scaling laws describe how model performance improves as a power law with size, but existing work has focused almost entirely on models above 100M parameters. The regime below 20 million parameters---where TinyML and edge AI systems operate---remains largely unexamined.
We train 90 models spanning 22K to 19.8M parameters across two architecture families (a plain ConvNet and MobileNetV2) on CIFAR-100, varying width while holding depth and training protocol fixed.
Both architectures follow approximate power laws in error rate, with exponents of $\alpha = 0.156 \pm 0.002$ (ScaleCNN) and $\alpha = 0.106 \pm 0.001$ (MobileNetV2), computed across five independent seeds. Because prior scaling laws were fit to cross-entropy loss rather than error rate, and these metrics are nonlinearly related, direct numerical comparison of exponents across regimes is approximate; with that caveat, the small-regime exponents are 1.4--2$\times$ steeper than the $\alpha \approx 0.076$ reported for large language models.
However, the power law does not hold uniformly: local exponents decay with scale, and MobileNetV2 saturates at 19.8M parameters ($\alpha_{\mathrm{local}} = 0.006$), reaching a capacity ceiling on this dataset.
The structure of errors also changes with scale. The Jaccard overlap between error sets of the smallest and largest ScaleCNN models is only 0.35 (mean across 25 seed pairs, $\pm 0.004$)---compression changes \emph{which} inputs are misclassified, not merely how many. Small models develop a triage strategy, concentrating capacity on easy classes (Gini of per-class accuracy: 0.26 at 22K params vs.\ 0.09 at 4.7M) while effectively abandoning the hardest ones (bottom-5 class accuracy: 10\% vs.\ 53\%). Counter to the usual trend of overconfidence scaling with capacity, the smallest models are the best calibrated (ECE = 0.013 vs.\ peak 0.110 at mid-size).
Aggregate accuracy alone is therefore a misleading basis for edge deployment decisions; validation must happen at the target model size.
\end{abstract}


\vspace{1em}

\section{Introduction}\label{sec:intro}

Neural scaling laws~\citep{kaplan2020scaling, hoffmann2022training} have become one of the more reliable empirical regularities in deep learning. Loss decreases as a power law with model size $N$, dataset size $D$, and compute budget $C$:
\begin{equation}
    L(N) \sim N^{-\alpha}
    \label{eq:scaling_law}
\end{equation}
where $\alpha \approx 0.076$ for language models (measured on cross-entropy loss), with similar values reported for vision transformers~\citep{zhai2022scaling}. These laws, spanning seven or more orders of magnitude, enabled the strategic scaling decisions behind GPT-3, PaLM, and subsequent frontier models.

Much less is known about the \emph{left side} of the scaling curve. Below 20 million parameters---the operating point of TinyML systems deployed on microcontrollers with $\leq$256\,KB RAM and $\leq$1\,mW power budgets---we lack answers to basic questions:

\begin{itemize}
    \item Does the same power law extend to the small-model regime, or does a different scaling relationship govern?
    \item Does compression merely increase the error rate, or does it fundamentally \emph{reshape} which inputs the model fails on?
    \item How do calibration and per-class fairness vary with scale?
\end{itemize}

These questions matter for deployment. Edge AI systems run in safety-critical contexts---autonomous vehicles, medical devices, industrial monitoring---where the failure \emph{distribution} matters as much as the failure \emph{rate}. A model that maintains 88\% accuracy after compression but silently shifts all its errors to a specific subpopulation may be more dangerous than one with 85\% accuracy and uniformly distributed errors.

The TinyML community typically reports model accuracy at individual operating points. To our knowledge, no prior work has systematically varied model size \emph{below 1M parameters} across orders of magnitude on a fixed task to characterize both the functional form and the qualitative changes in error behavior. We set out to do exactly that.

\paragraph{Contributions.} This paper makes three contributions:
\begin{enumerate}
    \item \textbf{Systematic scaling law characterization in the sub-20M regime.} We measure the accuracy--model-size relationship across nearly three orders of magnitude (22K--19.8M parameters) with two architecture families on CIFAR-100. We report architecture-dependent exponents that are 1.4--2$\times$ steeper than the large-model regime, though this comparison is approximate because prior work fit to cross-entropy loss while we fit to error rate (Section~\ref{sec:results_scaling}). We also document a broken power law and architecture-specific saturation.
    \item \textbf{Error redistribution under width scaling.} The Jaccard overlap between error sets of the smallest and largest models is only 0.35 ($\pm 0.004$ across 25 seed pairs), below the 0.42 expected from pure subset containment and well above the 0.21 expected from independent errors. Compression changes \emph{which} inputs are misclassified, not merely how many.
    \item \textbf{Class triage and calibration inversion.} Small models develop extreme triage strategies (Gini coefficient of per-class accuracy: 0.26 at 22K parameters vs.\ 0.09 at 4.7M), and the smallest models are the best calibrated, with ECE following an inverted-U pattern in ScaleCNN---contrary to the usual assumption that overconfidence grows monotonically with capacity.
\end{enumerate}

\section{Related Work}\label{sec:related}

\paragraph{Neural scaling laws.}
Kaplan et al.~\citep{kaplan2020scaling} established that language model loss follows power laws in model size, dataset size, and compute, with smooth trends spanning seven orders of magnitude. Hoffmann et al.~\citep{hoffmann2022training} refined these findings by characterizing compute-optimal training (Chinchilla scaling). Zhai et al.~\citep{zhai2022scaling} extended scaling analysis to vision transformers. Clark et al.~\citep{clark2022unified} proposed unified scaling laws across modalities. Hestness et al.~\citep{hestness2017deep} demonstrated that deep learning scaling is predictable across domains, reporting exponents in the range $\alpha \approx 0.07$--$0.12$ for vision tasks above 1M parameters. Rosenfeld et al.~\citep{rosenfeld2020constructive} proposed constructive methods for predicting generalization error across scales, including relatively small models. However, all of these works focus primarily on the large-model regime ($>$10M parameters) or treat smaller models as interpolation points rather than as a regime of independent interest. The sub-1M regime relevant to edge deployment has received little attention.

Sharma and Kaplan~\citep{sharma2022scaling} provided a theoretical grounding for scaling laws by relating the exponent to the intrinsic dimension of the data manifold, predicting $\alpha \approx 4/d$ where $d$ is the manifold dimension. Bahri et al.~\citep{bahri2024explaining} offered complementary explanations rooted in statistical mechanics. Caballero et al.~\citep{caballero2023broken} introduced \emph{broken neural scaling laws} (BNSL), showing that scaling behavior across wide parameter ranges is better described by smoothly bending power laws than by a single exponent. Our empirical observation of decaying local exponents is consistent with the BNSL framework.

\paragraph{TinyML and efficient inference.}
The TinyML community has developed a rich toolkit for deploying neural networks on microcontrollers: knowledge distillation~\citep{hinton2015distilling}, structured pruning~\citep{li2017pruning}, quantization-aware training~\citep{jacob2018quantization}, and neural architecture search for constrained devices~\citep{lin2020mcunet, banbury2021micronets}. These works optimize for accuracy at a \emph{specific} target size but do not characterize the functional relationship between accuracy and size across scales. We aim to fill that gap by measuring how accuracy, error distribution, and calibration evolve as model size varies continuously.

\paragraph{Emergent capabilities and phase transitions.}
Wei et al.~\citep{wei2022emergent} documented emergent capabilities in large language models that appear discontinuously at critical scales. Schaeffer et al.~\citep{schaeffer2023emergent} argued that some apparent emergence is an artifact of nonlinear evaluation metrics. Our work addresses the \emph{opposite} end of the scale: we study capability degradation as models shrink, focusing on per-class performance, error redistribution, and calibration rather than aggregate metrics.

\paragraph{Compression and fairness.}
Hooker et al.~\citep{hooker2020characterising} showed that pruning disproportionately affects underrepresented classes, introducing the concept of ``Pruning Identified Exemplars.'' Liebenwein et al.~\citep{liebenwein2021lost} extended this to show that pruning effects go beyond test accuracy. A key difference in experimental design: both Hooker et al.\ and Liebenwein et al.\ study a \emph{single trained model} subjected to post-hoc compression (pruning), whereas we train \emph{separate models from scratch} at each size. This means our error redistribution results reflect capacity-driven differences in what networks learn, not artifacts of which weights a pruning algorithm removes. Our work further extends these findings by (a)~quantifying the error redistribution across the full scaling range using Jaccard overlap rather than individual exemplars, (b)~demonstrating that scale drives error patterns more than architecture choice at matched parameter counts, and (c)~characterizing the class-level triage strategy that small models adopt.

\paragraph{Calibration.}
Guo et al.~\citep{guo2017calibration} demonstrated that modern neural networks are systematically miscalibrated, with overconfidence increasing with model capacity. Our results partially contradict this finding: in the small-model regime, the smallest models are the \emph{best} calibrated, and miscalibration follows an inverted-U pattern rather than monotonically increasing with size. The inverted-U pattern is reminiscent of the ``double descent'' phenomenon documented by Nakkiran et al.~\citep{nakkiran2021deep}, where test error can increase before decreasing as model size grows; our ECE results suggest a similar non-monotonicity in calibration. Alabdulmohsin et al.~\citep{alabdulmohsin2022revisiting} revisited scaling laws for vision and language with a focus on metrics beyond loss, which is close in spirit to our multi-metric approach.

\section{Theoretical Framework: Spectral Capacity Theory}\label{sec:theory}

This section applies and extends existing spectral analysis frameworks---particularly Sharma and Kaplan's~\citep{sharma2022scaling} data-manifold-dimension theory---to derive a prediction for the scaling exponent in the small-model regime. The core relation $\alpha = \gamma(\beta - 1)$ linking the scaling exponent to the data spectral decay ($\beta$) and the architecture's rank efficiency ($\gamma$) follows from their framework; our contribution is (a)~measuring $\beta$ directly from the CIFAR-100 eigenspectrum rather than using generic natural-image estimates, and (b)~showing that the implied $\gamma$ values differ between architectures in a way that explains the observed exponent gap.

\subsection{Data Spectral Structure}\label{sec:theory_spectral}

Natural data distributions exhibit scale-free spectral structure. The eigenvalues of the data covariance matrix $\Sigma$ follow a power-law decay:
\begin{equation}
    \lambda_k \sim k^{-\beta}
    \label{eq:spectral_decay}
\end{equation}
where $\beta > 1$ is the \emph{spectral decay exponent} and $k$ indexes eigenvalues in decreasing order. For natural images, $\beta \approx 1.0$--$1.2$~\citep{ruderman1994statistics, hyvarinen2009natural}. This power-law structure is a consequence of hierarchical spatial correlations in natural images.

\subsection{Effective Rank and the Parameter-to-Capacity Mapping}\label{sec:theory_rank}

A neural network with $N$ parameters can effectively represent data projected onto a subspace of dimensionality $K(N)$, which we term the \emph{effective rank}. For convolutional networks with width multiplier $m$:
\begin{equation}
    N \sim m^2 \cdot (\text{kernel size})^2 \cdot d, \quad K \sim m \cdot r
    \label{eq:param_rank}
\end{equation}
where $d$ is depth and $r$ is spatial resolution. In general:
\begin{equation}
    K(N) \sim N^{\gamma}
    \label{eq:rank_scaling}
\end{equation}
where $\gamma \in (0, 1)$ is the \emph{rank efficiency exponent}. For convolutional networks with quadratic parameter scaling ($N \propto m^2$), the effective rank scales linearly with width ($K \propto m \propto N^{0.5}$), giving $\gamma \approx 0.5$.

\subsection{Deriving the Scaling Exponent}\label{sec:theory_exponent}

A model with effective rank $K(N)$ captures the top $K(N)$ eigenmodes of the data distribution. The residual loss is:
\begin{equation}
    L(N) = \sum_{k > K(N)} \lambda_k \approx \int_{K(N)}^{\infty} k^{-\beta} \, dk = \frac{K(N)^{1-\beta}}{\beta - 1}
    \label{eq:residual_loss}
\end{equation}
Substituting $K(N) \sim N^{\gamma}$:
\begin{equation}
    \boxed{L(N) \sim N^{-\alpha}, \quad \text{where} \quad \alpha = \gamma(\beta - 1)}
    \label{eq:exponent_prediction}
\end{equation}
This predicts the scaling exponent from two independently measurable quantities: $\beta$ (a data property) and $\gamma$ (an architecture property).

\subsection{Reconciling Theory and Measurement}\label{sec:theory_reconcile}

For natural images with $\beta \approx 1.1$ and convolutional networks with $\gamma \approx 0.5$, the spectral capacity theory predicts $\alpha \approx 0.05$. Our measured exponents---$\alpha = 0.156$ for ScaleCNN and $\alpha = 0.106$ for MobileNetV2---are 2--3$\times$ larger.

To test the theory quantitatively, we computed $\beta$ directly from the CIFAR-100 training set by performing PCA on the 50,000 $\times$ 3,072 data matrix and fitting $\lambda_k \sim k^{-\beta}$ to eigenvalues $k = 10$ through 500 (skipping the dominant modes, which carry most variance but do not follow the power law). The result is $\beta_{\mathrm{CIFAR}} = 1.45 \pm 0.004$ ($R^2 = 0.996$), substantially steeper than the $\beta \approx 1.1$ reported for natural images at native resolution~\citep{ruderman1994statistics}. The steeper decay likely reflects CIFAR-100's low resolution (32$\times$32) and fine-grained class structure, which concentrate information in fewer modes.

With $\beta = 1.45$, the theory predicts $\alpha = \gamma \times 0.45$. Using the naive estimate $\gamma = 0.5$ yields $\alpha = 0.225$, which \emph{overpredicts} both measured exponents. Inverting the relation gives the implied rank efficiency:
\begin{equation}
    \gamma_{\mathrm{CNN}} = \frac{\alpha_{\mathrm{CNN}}}{\beta - 1} = \frac{0.156}{0.45} \approx 0.35, \quad \gamma_{\mathrm{MobNet}} = \frac{0.106}{0.45} \approx 0.24
    \label{eq:implied_gamma}
\end{equation}
Both values are below the naive $\gamma = 0.5$, indicating that not all parameters contribute equally to spectral capacity. The gap between architectures ($\gamma_{\mathrm{CNN}} / \gamma_{\mathrm{MobNet}} \approx 1.5$) is consistent with MobileNetV2's depthwise separable design spending parameters on bottleneck projections and expansion layers that do not proportionally increase effective rank at small widths.

The theory correctly predicts the functional form (power law with architecture-dependent exponent), the qualitative ordering ($\alpha_{\mathrm{CNN}} > \alpha_{\mathrm{MobNet}}$), and---with the measured $\beta$---gives quantitative estimates of the rank efficiency. Measuring $\gamma$ directly from the effective rank of trained networks would close the loop, which we leave to future work.

\section{Experimental Setup}\label{sec:experiments}

\subsection{Hypotheses}\label{sec:exp_hypotheses}

We organize the experiments around two hypotheses.

\textbf{H1 (Power-law scaling).} If the sub-20M regime follows the same scaling form as larger models, then $\log(\text{error rate})$ vs.\ $\log(N)$ should be approximately linear across $\geq$2 orders of magnitude. We consider the hypothesis unsupported if $R^2 < 0.9$ or if residuals show systematic non-monotonic structure.

\textbf{H2 (Error redistribution).} If compression merely adds errors without changing which inputs fail, then the Jaccard overlap between error sets at different model sizes should remain high. We consider the hypothesis unsupported if overlap stays above 0.90 across all comparisons.

\subsection{Model Families}\label{sec:exp_models}

We use two architecture families to span complementary parameter ranges:

\paragraph{ScaleCNN.} A plain 4-block convolutional network with channels $[c, 2c, 4c, 8c]$, where each block contains two conv-BN-ReLU layers followed by spatial pooling. The classifier is a single linear layer. Base channel widths $c \in \{4, 8, 12, 16, 24, 32, 48, 64\}$ yield 8 model sizes from 22K to 4.7M parameters with clean quadratic scaling ($N \propto c^2$). This architecture was chosen for its interpretable capacity scaling and absence of structural bottlenecks.

\paragraph{MobileNetV2.} The standard inverted-residual architecture~\citep{sandler2018mobilenetv2} with width multipliers $m \in \{0.10, 0.15, 0.25, 0.35, 0.50, 0.75, 1.00, 1.50, 2.00, 3.00\}$, yielding 10 model sizes from 214K to 19.8M parameters. The first convolutional layer's stride is reduced from 2 to 1 to accommodate CIFAR-100's 32$\times$32 input resolution.

Together, the two families provide 18 configurations spanning 22K to 19.8M parameters (nearly three orders of magnitude). Table~\ref{tab:configs} summarizes the experimental grid.

\begin{table}[H]
\centering
\caption{Experimental configurations: 18 model sizes $\times$ 5 seeds = 90 runs.}
\label{tab:configs}
\begin{tabular}{llcc}
\toprule
\textbf{Architecture} & \textbf{Size Parameter} & \textbf{Configs} & \textbf{Param Range} \\
\midrule
ScaleCNN & Base channels $c \in \{4, 8, ..., 64\}$ & 8 & 22K -- 4.7M \\
MobileNetV2 & Width mult.\ $m \in \{0.10, ..., 3.00\}$ & 10 & 214K -- 19.8M \\
\midrule
\multicolumn{2}{l}{\textbf{Total}} & \textbf{18} & \textbf{22K -- 19.8M} \\
\bottomrule
\end{tabular}
\end{table}

\subsection{Dataset: CIFAR-100}\label{sec:exp_data}

We evaluate on CIFAR-100, which provides 50,000 training images and 10,000 test images across 100 balanced classes at 32$\times$32 resolution. CIFAR-100 was chosen because (a)~its 100 fine-grained classes create meaningful per-class variation, enabling the class triage analysis; (b)~its moderate training set size creates natural saturation behavior at the upper end of our parameter range; and (c)~it is a widely used benchmark in the TinyML and model compression literature.

\subsection{Training Protocol}\label{sec:exp_training}

All models within an architecture family share identical training hyperparameters: SGD with momentum 0.9, weight decay $5 \times 10^{-4}$, cosine annealing from $10^{-1}$ to $10^{-5}$ over 200 epochs, batch size 128, and data augmentation (random crop with padding 4, horizontal flip, Cutout~\citep{devries2017improved} with 8$\times$8 patches). Each configuration is trained with \textbf{five random seeds} (0--4). All results report mean $\pm$ standard deviation across seeds.

\subsection{Measurements}\label{sec:exp_measurements}

At each operating point, we measure:
\begin{itemize}
    \item \textbf{Performance:} Top-1 accuracy, top-5 accuracy, per-class accuracy vector (100 values).
    \item \textbf{Error characterization:} Per-sample correct/incorrect mask (for Jaccard overlap computation), per-class error rates, Gini coefficient of per-class accuracy.
    \item \textbf{Calibration:} Expected calibration error (ECE, 15 bins), mean confidence on correct and incorrect predictions.
\end{itemize}

Evaluation uses the trained model at epoch 200 (no early stopping or best-checkpoint selection) to ensure all models are compared at the same training budget.

\section{Results}\label{sec:results}

\subsection{Power-Law Scaling}\label{sec:results_scaling}

\begin{figure}[H]
    \centering
    \includegraphics[width=0.95\textwidth]{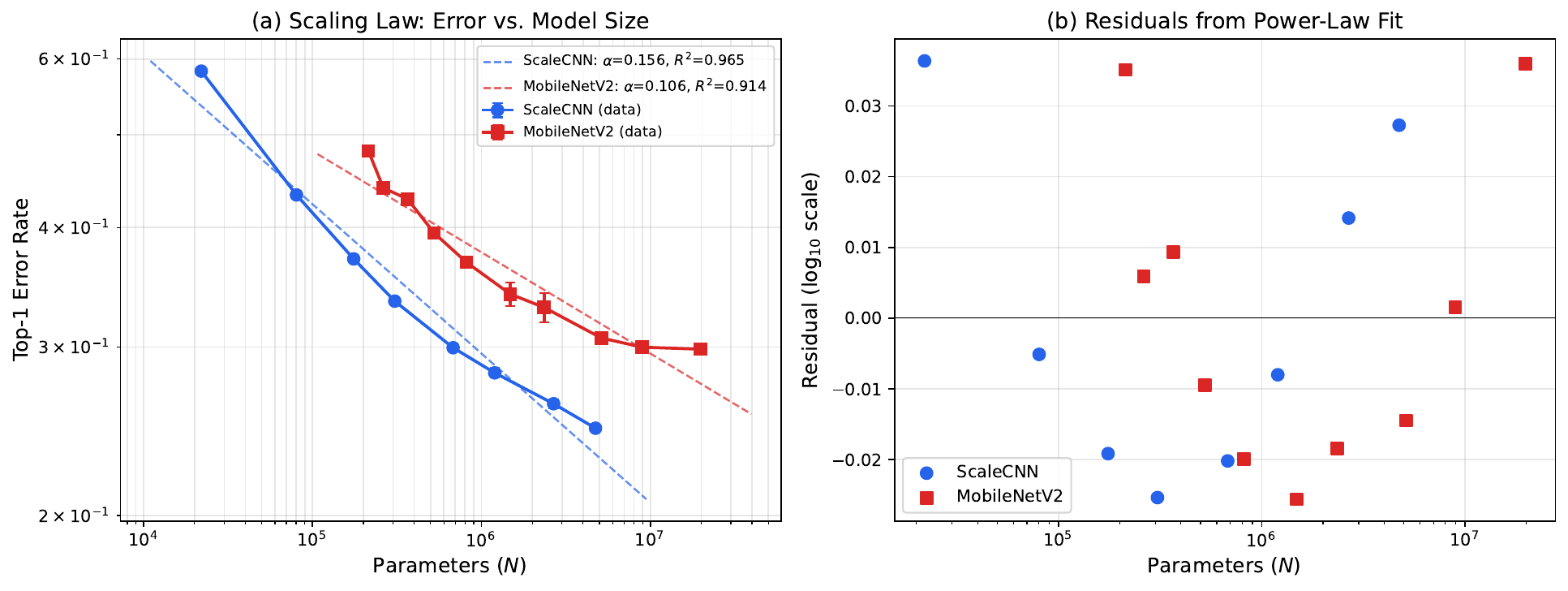}
    \caption{(a) Log-log plot of top-1 error rate vs.\ parameter count for ScaleCNN (blue) and MobileNetV2 (red). Error bars: $\pm 1$ std across 5 seeds. Dashed lines: power-law fits. (b) Residuals from the power-law fit showing systematic curvature (broken power law).}
    \label{fig:scaling_law}
\end{figure}

Figure~\ref{fig:scaling_law} plots top-1 error rate vs.\ parameter count on log-log axes for both architecture families. Both exhibit approximate power-law behavior:
\begin{align}
    \text{ScaleCNN:} \quad & \text{error} \sim N^{-0.156 \pm 0.002}, \quad R^2 = 0.965 \label{eq:cnn_fit} \\
    \text{MobileNetV2:} \quad & \text{error} \sim N^{-0.106 \pm 0.001}, \quad R^2 = 0.914 \label{eq:mob_fit}
\end{align}
where the $\pm$ values are standard deviations across five independent seeds (95\% CIs: $[0.154, 0.158]$ and $[0.105, 0.107]$, respectively). The difference between architectures is highly significant ($t = 50.9$, $p < 10^{-6}$, two-sample $t$-test on per-seed exponents). The MobileNetV2 fit is weaker ($R^2 = 0.914$, per-seed range: 0.860--0.928) than ScaleCNN's ($R^2 = 0.965$, per-seed range: 0.961--0.970), likely reflecting the oscillatory local exponents and saturation behavior discussed in Sections~\ref{sec:results_broken}--\ref{sec:results_saturation}.

ScaleCNN exhibits a 47\% steeper scaling exponent than MobileNetV2, despite the latter being engineered for inference efficiency. We attribute the gap to MobileNetV2's structural overhead: depthwise separable convolutions, expansion layers, and bottleneck projections consume parameters without proportionally increasing representational capacity at small widths.

Both exponents are markedly steeper than the $\alpha \approx 0.076$ reported for large language models~\citep{kaplan2020scaling} and the $\alpha \approx 0.07$--$0.12$ range reported for vision models above 1M parameters~\citep{hestness2017deep}. However, a caveat is warranted: Kaplan et al.\ and Hestness et al.\ fit power laws to \emph{cross-entropy loss}, while we fit to \emph{error rate} ($1 - \text{accuracy}$). Because error rate is a nonlinear, bounded transformation of loss, power-law exponents in the two metrics are not directly comparable; the steeper exponents we observe may partly reflect the metric difference rather than a genuine regime change.

As partial evidence for the magnitude of the metric effect, we fit power laws to final training cross-entropy loss (available for all 90 runs): the training-loss exponents are $\alpha_{\mathrm{train}} = 0.84 \pm 0.004$ (ScaleCNN) and $0.30 \pm 0.003$ (MobileNetV2)---5$\times$ and 3$\times$ steeper than the corresponding error-rate exponents. Training loss is not a substitute for test loss (the largest models are heavily overfit, with training loss of 0.028 vs.\ 24.7\% test error), but the large difference confirms that loss and error rate scale very differently in this regime, and direct exponent comparison across metrics is unreliable. We also note that with only 8 (ScaleCNN) and 10 (MobileNetV2) model sizes, the $R^2$ values do not discriminate between a power law and other smooth functions (e.g., log-linear, BNSL); formal model selection via AIC/BIC with denser sampling would be needed to settle the functional form (see Section~\ref{sec:limitations}).

Table~\ref{tab:main_results} provides the full results.

\begin{table}[H]
\centering
\caption{Full results: 90 runs (5 seeds each). Accuracy is top-1 on CIFAR-100 test set.}
\label{tab:main_results}
\begin{tabular}{llrrrr}
\toprule
\textbf{Arch} & \textbf{Config} & \textbf{Params} & \textbf{MACs (M)} & \textbf{Accuracy (\%)} & \textbf{ECE} \\
\midrule
ScaleCNN & $c=4$   &    21,936 &   1.8 & $41.71 \pm 0.45$ & 0.013 \\
         & $c=8$   &    80,348 &   7.0 & $56.73 \pm 0.40$ & 0.037 \\
         & $c=12$  &   175,336 &  15.5 & $62.91 \pm 0.17$ & 0.059 \\
         & $c=16$  &   306,900 &  27.4 & $66.49 \pm 0.23$ & 0.073 \\
         & $c=24$  &   679,756 &  61.2 & $70.05 \pm 0.29$ & 0.098 \\
         & $c=32$  & 1,198,916 & 108.6 & $71.81 \pm 0.33$ & 0.110 \\
         & $c=48$  & 2,676,148 & 243.8 & $73.82 \pm 0.28$ & 0.102 \\
         & $c=64$  & 4,738,596 & 433.0 & $75.32 \pm 0.22$ & 0.082 \\
\midrule
MobNetV2 & $m=0.10$ &   214,180 &  5.3 & $51.91 \pm 0.42$ & 0.016 \\
         & $m=0.15$ &   262,596 &  8.1 & $55.99 \pm 0.37$ & 0.021 \\
         & $m=0.25$ &   366,212 & 14.5 & $57.17 \pm 0.38$ & 0.023 \\
         & $m=0.35$ &   524,228 & 22.8 & $60.52 \pm 0.52$ & 0.033 \\
         & $m=0.50$ &   815,780 & 40.1 & $63.21 \pm 0.34$ & 0.038 \\
         & $m=0.75$ & 1,483,524 & 78.6 & $65.92 \pm 0.98$ & 0.045 \\
         & $m=1.00$ & 2,351,972 & 131.4 & $67.00 \pm 1.16$ & 0.053 \\
         & $m=1.50$ & 5,129,252 & 280.3 & $69.34 \pm 0.23$ & 0.058 \\
         & $m=2.00$ & 8,953,188 & 490.7 & $70.01 \pm 0.12$ & 0.059 \\
         & $m=3.00$ &19,803,748 &1089.2 & $70.15 \pm 0.28$ & 0.072 \\
\bottomrule
\end{tabular}
\end{table}

As an illustrative comparison, ScaleCNN at $c=64$ (4.7M parameters, 75.32\% accuracy) outperforms MobileNetV2 at $m=3.0$ (19.8M parameters, 70.15\% accuracy) by over 5 percentage points with 4.2$\times$ fewer parameters. This gap is partly inflated by comparing against MobileNetV2's saturation endpoint (Section~\ref{sec:results_saturation}); a fairer comparison at matched parameter counts ($\sim$2.5M) still favors ScaleCNN by $\sim$4 points (73.8\% vs.\ 67.0\%). At this dataset scale, the plain convolutional architecture achieves higher parameter efficiency than MobileNetV2, which was designed for \emph{inference} efficiency (low FLOPs and latency) rather than parameter efficiency.

\subsection{Broken Power Law: Local Exponent Decay}\label{sec:results_broken}

\begin{figure}[H]
    \centering
    \includegraphics[width=0.7\textwidth]{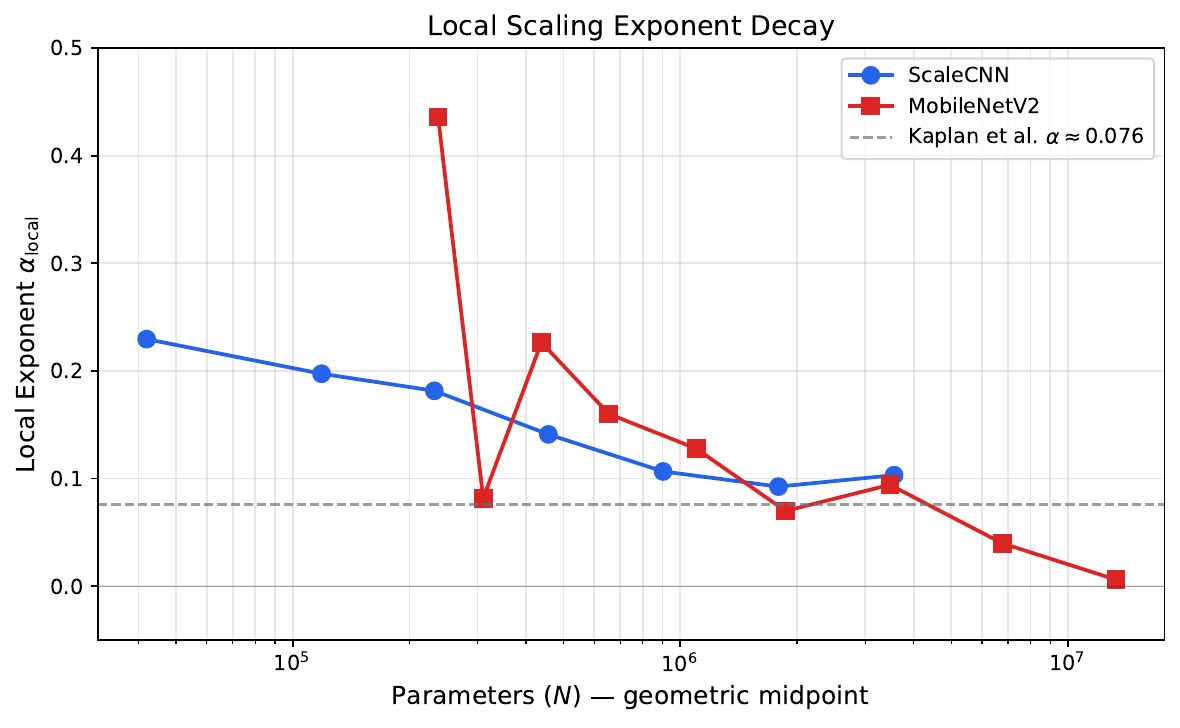}
    \caption{Local scaling exponent $\alpha_{\mathrm{local}}$ vs.\ parameter count. ScaleCNN shows smooth monotonic decay from 0.23 to 0.10. MobileNetV2 shows oscillatory behavior followed by collapse to 0.006. The Kaplan et al.\ reference exponent ($\alpha \approx 0.076$) is shown as a dashed line.}
    \label{fig:local_exponents}
\end{figure}

A single exponent summarizes the full range but hides structure. Figure~\ref{fig:local_exponents} shows what happens when we compute the local scaling exponent $\alpha_{\mathrm{local}}$ as the finite-difference slope between adjacent model sizes: it decays with scale.

ScaleCNN exhibits smooth, monotonic decay: the local exponent falls from 0.23 in the tiny regime ($<$100K parameters) to 0.10 near 5M parameters, approaching the range reported for larger vision models. This is consistent with a smoothly bending power law (the BNSL framework of Caballero et al.~\citep{caballero2023broken}) rather than a sharp regime change.

MobileNetV2 behaves differently. The local exponent oscillates wildly at small widths---jumping from 0.44 ($m=0.10 \to 0.15$) down to 0.08 ($m=0.15 \to 0.25$) and back up to 0.23 ($m=0.25 \to 0.35$). We initially thought this was a seed variance artifact and reran the smallest MobileNetV2 configurations, but the pattern was stable across all five seeds. We do not have a definitive explanation for this oscillation. One hypothesis is that it reflects structural bottlenecks in the inverted-residual design---specifically, that certain width multipliers cause the bottleneck projection dimension to cross integer thresholds that discretely change the network's representational capacity. Testing this hypothesis would require ablating individual layers at each width, which we leave to future work. Above $m=0.50$, the exponent declines monotonically toward zero.

In the tiny-model regime ($<$100K parameters), local exponents reach 0.23, suggestive of steeper scaling than the $\alpha \approx 0.07$--$0.12$ reported for larger models, though this observation rests on only 2--3 data points and should be treated as indicative rather than precise.

\subsection{MobileNetV2 Capacity Saturation}\label{sec:results_saturation}

The final MobileNetV2 data points reveal complete architectural saturation on CIFAR-100:

\begin{center}
\begin{tabular}{lrrl}
\toprule
\textbf{Transition} & \textbf{Accuracy Gain} & \textbf{$\alpha_{\mathrm{local}}$} & \\
\midrule
$m=1.0 \to 1.5$ & $+2.34\%$ & 0.094 & \\
$m=1.5 \to 2.0$ & $+0.67\%$ & 0.040 & Diminishing returns \\
$m=2.0 \to 3.0$ & $+0.14\%$ & 0.006 & Saturation \\
\bottomrule
\end{tabular}
\end{center}

Doubling from 9M to 20M parameters yields only 0.14\% accuracy gain ($\alpha_{\mathrm{local}} = 0.006$), effectively zero. At $m=3.0$, the parameter-to-sample ratio is 396:1, and the model is massively overparameterized for CIFAR-100's 50K training set. This saturation is \emph{not} observed in ScaleCNN at its largest tested size ($c=48 \to 64$ still shows $\alpha_{\mathrm{local}} = 0.10$ at 4.7M parameters), suggesting the saturation is architecture-specific. However, ScaleCNN was only tested up to 4.7M parameters vs.\ MobileNetV2's 19.8M, so we cannot rule out that ScaleCNN would also saturate at higher widths. The comparison suggests that MobileNetV2's depthwise separable design hits a capacity ceiling earlier than a simple ConvNet on this dataset, but confirming this would require extending ScaleCNN to comparable parameter counts.

\subsection{Error Redistribution}\label{sec:results_error}

The scaling exponent tells us how much worse a smaller model gets. It says nothing about \emph{how} it gets worse---and in practice, the ``how'' matters more. Compression does not just add errors; it changes which inputs are misclassified.

\begin{figure}[H]
    \centering
    \includegraphics[width=0.7\textwidth]{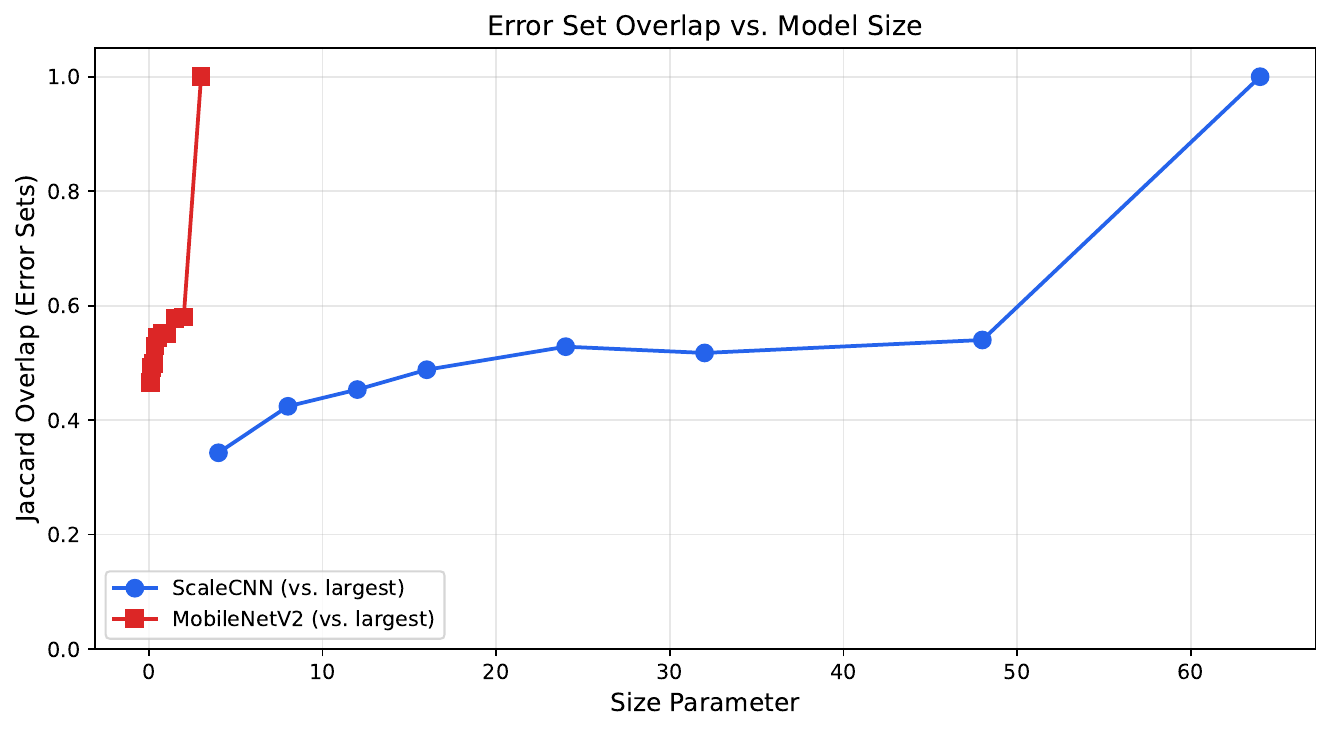}
    \caption{Jaccard overlap between error sets of each model and the largest model in its family. ScaleCNN $c=4$ vs.\ $c=64$: Jaccard = $0.35 \pm 0.004$ (25 seed pairs). Adjacent model sizes share 65--72\% of errors.}
    \label{fig:error_overlap}
\end{figure}

We compute the Jaccard index between error sets of each model and the largest model in its architecture family. Because each seed produces a different per-sample error mask, we compute Jaccard for all 25 seed pairs (5 seeds $\times$ 5 seeds) between model sizes and report the mean $\pm$ standard deviation. As a stability baseline, cross-seed Jaccard \emph{within} the same configuration (e.g., ScaleCNN $c=4$ seed-0 vs.\ seed-1) is $0.704 \pm 0.007$, confirming that roughly 70\% of errors are reproducible across random initializations at fixed model size. Figure~\ref{fig:error_overlap} shows the cross-scale results.

The ScaleCNN Jaccard matrix shows that only 35\% of errors overlap between the smallest ($c=4$, 22K params) and largest ($c=64$, 4.7M params) models ($J = 0.349 \pm 0.004$ across 25 seed pairs). A 200$\times$ compression changes the identity of 65\% of errors. For comparison, the cross-seed Jaccard \emph{within} the $c=4$ configuration is 0.704---so roughly 30\% of errors change due to random initialization alone, while the additional 35 percentage points of error change between $c=4$ and $c=64$ reflect genuine scale-driven redistribution. Adjacent sizes share 65--72\% of errors, showing that error redistribution accumulates gradually across the scaling range.

To calibrate the magnitude of this effect, we compare against two reference models. If the smaller model's errors were a strict superset of the larger model's (maximal subset containment), the Jaccard would equal $e_{\mathrm{large}} / e_{\mathrm{small}} \approx 0.42$.\footnote{For ScaleCNN: $e_{c=64} = 0.247$, $e_{c=4} = 0.583$. Maximal containment Jaccard $= e_{\mathrm{large}} / e_{\mathrm{small}} = 0.247 / 0.583 = 0.42$.} If errors were statistically independent at the observed accuracy levels, the expected Jaccard would be:
\begin{equation}
    J_{\mathrm{indep}} = \frac{e_A \cdot e_B}{e_A + e_B - e_A \cdot e_B} = \frac{0.583 \times 0.247}{0.583 + 0.247 - 0.144} \approx 0.21
    \label{eq:jaccard_null}
\end{equation}
The observed value of 0.35 falls between these bounds: the two models' errors are positively correlated (both tend to fail on genuinely hard inputs), but there is substantial redistribution beyond what the accuracy difference alone would predict. The gap between the observed Jaccard (0.35) and the independence baseline (0.21) indicates real correlation; the gap between observed (0.35) and maximal containment (0.42) quantifies the redistribution effect. To assess statistical significance, we compute 95\% bootstrap confidence intervals (10,000 resamples of the 25 seed pairs): $J_{c=4 \text{ vs.\ } c=64} \in [0.342, 0.356]$, which excludes both the independence baseline (0.21) and the maximal containment value (0.42), confirming that the observed Jaccard is significantly different from both null models ($p < 0.001$ in each case).

Cross-architecture analysis shows that two different architectures at similar parameter counts make \emph{more similar} errors than the same architecture at different sizes. In our experiments, \textbf{scale drives error patterns more than architecture choice at matched parameter counts}, though this finding is based on two architecture families and would need broader validation.

The MobileNetV2 saturation effect is visible in the Jaccard data: $m=2.0$ and $m=3.0$ have nearly identical cross-architecture overlap values (e.g., vs.\ ScaleCNN $c=64$: 0.573 vs.\ 0.567), confirming that doubling parameters changes almost nothing about the error pattern once saturation is reached.

\subsection{Class Triage: The Fairness Cost of Compression}\label{sec:results_triage}

\begin{figure}[H]
    \centering
    \includegraphics[width=0.95\textwidth]{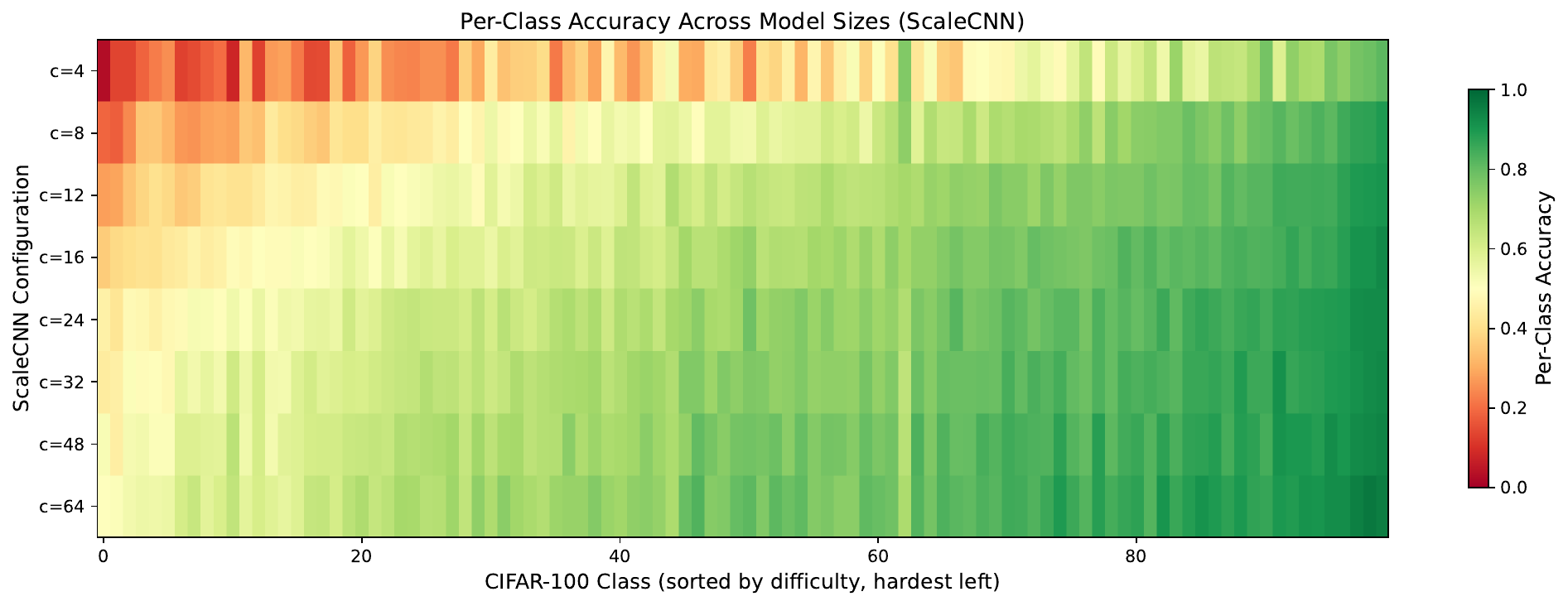}
    \caption{Per-class accuracy heatmap for ScaleCNN (classes sorted by difficulty, hardest at left). Color scale: dark red = low accuracy ($\leq$10\%), bright green = high accuracy ($\geq$80\%). The consistently hardest classes (leftmost columns) include visually similar fine-grained categories such as \emph{boy/girl}, \emph{maple/oak tree}, and \emph{leopard/tiger}. Small models show extreme variation: near-zero accuracy on these hardest classes, near-80\% on the easiest.}
    \label{fig:heatmap}
\end{figure}

\begin{figure}[H]
    \centering
    \includegraphics[width=0.95\textwidth]{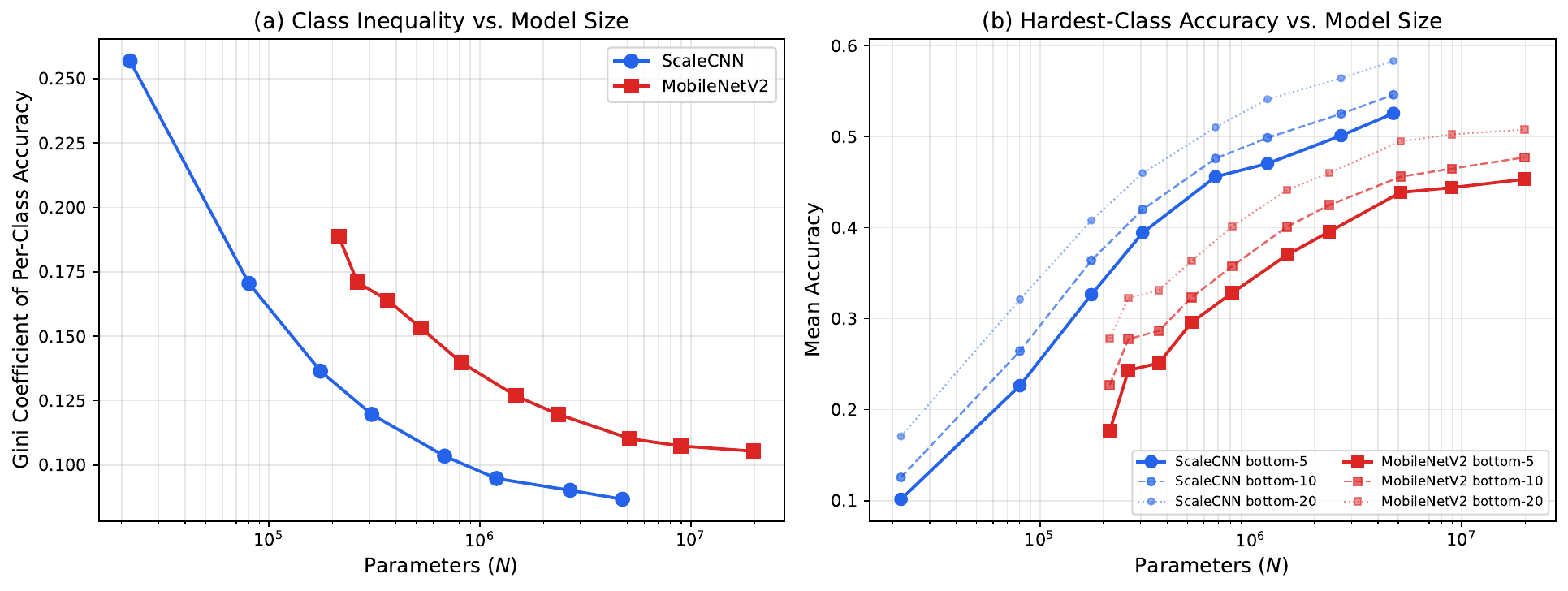}
    \caption{(a) Gini coefficient of per-class accuracy vs.\ model size. Small models exhibit high inequality (Gini = 0.26), which drops to 0.09 as models grow. (b) Mean accuracy of the hardest 5, 10, and 20 classes. Bottom-5 accuracy rises from 10\% to 53\% with scale.}
    \label{fig:class_triage}
\end{figure}

Small models develop an extreme triage strategy: they concentrate representational capacity on easy classes and all but abandon the hardest ones. Figures~\ref{fig:heatmap} and~\ref{fig:class_triage} quantify this effect.

The Gini coefficient drops 3$\times$ as models grow from 22K to 4.7M parameters (ScaleCNN: $0.257 \pm 0.011$ $\to$ $0.087 \pm 0.005$, mean $\pm$ std across 5 seeds; 95\% bootstrap CI for the difference: $[0.153, 0.187]$, $p < 0.001$), indicating that larger models distribute accuracy more uniformly across classes. To verify this is not a sampling artifact, we compute the expected Gini under a binomial null model where all 100 classes share the same true accuracy $p$ and each is evaluated on 100 test images. At $p = 0.42$ (the smallest ScaleCNN), the expected Gini from finite-sample variance alone is $\approx$0.067; the observed 0.257 is 3.8$\times$ larger, confirming that the triage effect reflects genuine class-level differences rather than statistical noise. At $p = 0.75$ (the largest ScaleCNN), the null Gini is $\approx$0.032 vs.\ observed 0.087 (2.7$\times$), so some per-class inequality persists even in larger models but is much closer to the sampling floor.

The gains from additional parameters are concentrated in the hardest classes: bottom-5 accuracy improves 5$\times$ (10.2\% $\to$ 52.6\%) while top-5 accuracy improves only 1.2$\times$ (78.3\% $\to$ 94.0\%).

MobileNetV2 shows the same pattern with saturation: $m=2.0 \to 3.0$ yields only 0.9\% gain in bottom-5 accuracy and negligible Gini change (0.107 $\to$ 0.105).

For deployment, this means that a model compressed to fit on a microcontroller will sacrifice rare-class performance first. If those rare classes correspond to safety-critical categories (e.g., unusual medical conditions, rare traffic scenarios), this degradation will not be captured by aggregate accuracy metrics.

\subsection{Calibration Inversion}\label{sec:results_calibration}

\begin{figure}[H]
    \centering
    \includegraphics[width=0.7\textwidth]{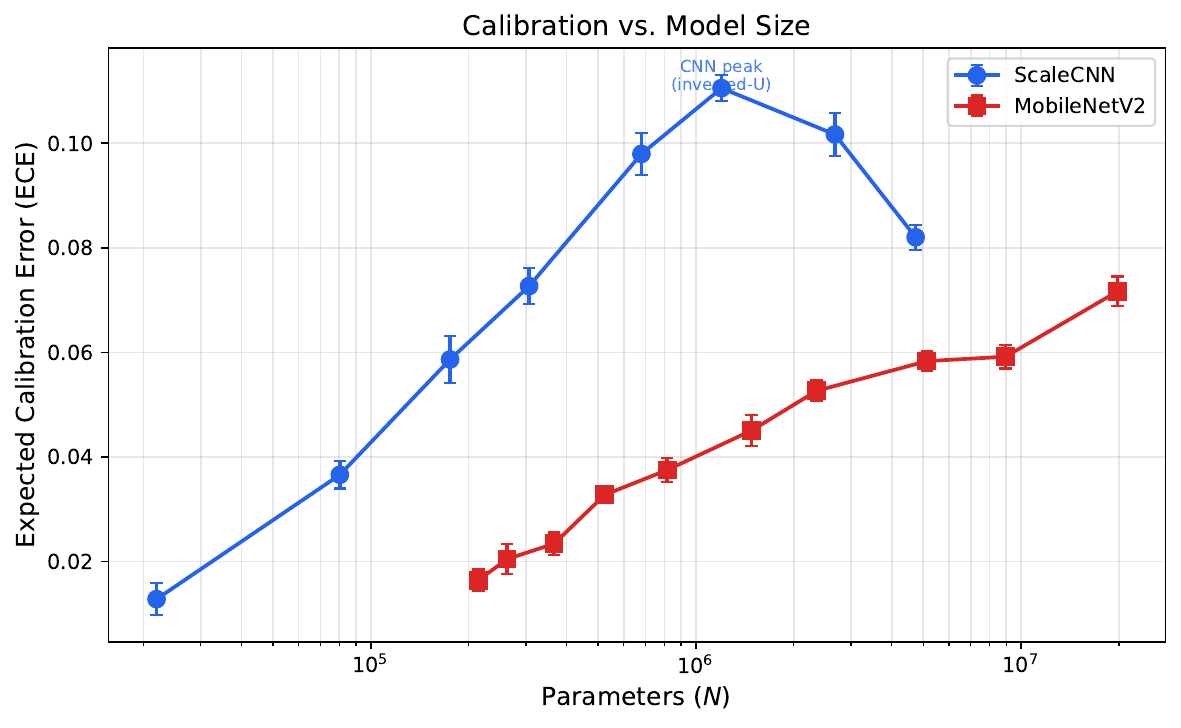}
    \caption{Expected Calibration Error (ECE) vs.\ model size. ScaleCNN shows an inverted-U pattern with peak miscalibration at $c=32$ (1.2M params). MobileNetV2 shows monotonically increasing ECE with a jump at the saturated $m=3.0$. Both architectures agree: the smallest models are the best calibrated.}
    \label{fig:calibration}
\end{figure}

Figure~\ref{fig:calibration} presents the relationship between ECE and model size.

The two architectures show different calibration trajectories:
\begin{itemize}
    \item \textbf{ScaleCNN}: An inverted-U pattern. ECE rises from 0.013 ($c=4$) to a peak of 0.110 ($c=32$, 1.2M parameters), then recovers to 0.082 ($c=64$, 4.7M parameters). Mid-sized models are the most miscalibrated.
    \item \textbf{MobileNetV2}: Monotonically increasing ECE from 0.016 ($m=0.10$) to 0.072 ($m=3.0$). At the saturated $m=3.0$ endpoint, ECE jumps upward even though accuracy barely changes---the model grows more overconfident without becoming more capable (cf.\ Section~\ref{sec:results_saturation}).
\end{itemize}

Despite their low accuracy, \textbf{the smallest models achieve the lowest ECE} in both families. A 22K-parameter ScaleCNN with 42\% accuracy (ECE = 0.013) produces lower ECE than a 1.2M-parameter model with 72\% accuracy (ECE = 0.110).

\paragraph{Important qualification: global vs.\ per-bin calibration.}
The mechanism behind the small model's low ECE warrants careful interpretation. Standard calibration requires sharp \emph{per-bin} alignment between predicted confidence and observed accuracy across all confidence levels. The $c=4$ model's low ECE instead reflects a near-match between its \emph{global} mean confidence (0.425) and overall accuracy (41.7\%), without necessarily achieving fine-grained bin-level reliability. In other words, a model that is uncertain about everything and wrong about half the time will appear well-calibrated by ECE, but for a different reason than a model that assigns high confidence to correct predictions and low confidence to incorrect ones.

Mid-sized models ($c=32$), by contrast, develop mean confidence of 0.83 despite 71.8\% accuracy, producing the overconfidence peak. The small model does carry some discriminative signal (mean confidence on correct predictions: 0.57 vs.\ incorrect: 0.32), so its confidence scores are not purely random---but the low ECE is primarily a global-match artifact rather than evidence of sharp per-sample calibration. In practice, confidence scores from a very small model may still be more useful for rejection or deferral decisions than those from a mid-sized overconfident one, though the mechanism is different from what ``well-calibrated'' typically implies.

\section{Discussion}\label{sec:discussion}

\subsection{Reconciling Theory and Measurement}\label{sec:discuss_theory}

The spectral capacity theory (Section~\ref{sec:theory}) correctly predicts the functional form and qualitative architecture dependence of the scaling exponent. Direct measurement of the CIFAR-100 eigenspectrum yields $\beta = 1.45$, substantially steeper than the $\beta \approx 1.1$ of natural images at native resolution. With this measured $\beta$, the theory implies rank efficiencies of $\gamma_{\mathrm{CNN}} \approx 0.35$ and $\gamma_{\mathrm{MobNet}} \approx 0.24$---both below the naive $\gamma = 0.5$ from width-counting arguments, indicating that parameter-to-capacity conversion is less efficient than assumed. Directly measuring $\gamma$ from the effective rank of trained networks (e.g., via the stable rank of activation matrices) would complete the quantitative validation.

\subsection{Architecture Dependence}\label{sec:discuss_arch}

The finding that $\alpha_{\mathrm{CNN}} > \alpha_{\mathrm{MobNet}}$ ($0.156$ vs.\ $0.106$) implies that simpler architectures scale more steeply in the small regime. MobileNetV2's depthwise separable design introduces structural overhead (1$\times$1 bottleneck projections, expansion layers) that consumes parameters without proportionally increasing representational capacity at small widths. This overhead washes out at large widths---MobileNetV2 was designed for efficient inference on ImageNet-scale models---but it becomes a significant capacity penalty in the sub-1M regime.

For very small parameter budgets ($<$500K), TinyML practitioners may be better served by plain convolutional architectures, reserving inference-optimized designs for larger models where their computational efficiency advantage outweighs the capacity penalty.

\subsection{Implications for Edge Deployment}\label{sec:discuss_deployment}

A common approach in edge deployment is to train a large model, compress it, verify that aggregate accuracy remains acceptable, and deploy. Our error redistribution results suggest this workflow is insufficient. With Jaccard overlap as low as 0.35 between the smallest and largest models, the error behavior of a large model is not predictive of which inputs the compressed model will misclassify. Validation must therefore occur at the target deployment size.

The cross-architecture comparison tells the same story: a different-architecture model at the target size provides a better estimate of deployment-time failure modes than the same architecture at a larger size. The MobileNetV2 saturation data adds a related caution---scaling beyond $m=1.5$ on CIFAR-100 yields negligible accuracy gains while increasing ECE, so model capacity should be matched to dataset size.

\subsection{Class-Level Disparities Under Capacity Constraints}\label{sec:discuss_fairness}

The Gini coefficient data reveals a quantitative ``fairness tax'' of compression. At 22K parameters, the per-class accuracy Gini is 0.257---meaning a substantial fraction of classes are largely neglected. As the model grows to 4.7M parameters, the Gini drops to 0.087, and accuracy gains flow disproportionately to the hardest classes.

The concern is sharpest in domains where rare classes carry outsized importance---uncommon conditions in medical imaging, unusual road scenarios in autonomous driving. Our data shows that small models degrade most on exactly these categories. The mechanism is straightforward: gradient descent on the empirical loss allocates capacity roughly in proportion to class learnability, so rare or visually ambiguous classes receive the least capacity and achieve the lowest accuracy when total capacity is constrained.

\section{Limitations}\label{sec:limitations}

The clearest limitation is scope: all results come from a single dataset (CIFAR-100, 32$\times$32, 50K training images, 100 balanced classes) and two architectures. The largest model has a parameter-to-sample ratio of 396:1, so the upper end of our range is partially a study of overfitting rather than pure capacity scaling. The TinyML architecture zoo is much larger than what we test---binary networks, NAS-derived architectures, and tiny vision transformers may behave differently---and our findings may not transfer to higher-resolution datasets or other modalities like audio.

A related concern is that we scale by width only. Depth is held fixed (4 blocks for ScaleCNN, 17 for MobileNetV2), so parameters, FLOPs, and activation memory are perfectly correlated in our experiments. We cannot disentangle capacity effects from compute effects. On the training side, our convergence criterion (loss change $<$1\% in the last 20 epochs) flagged most runs as not fully converged, likely an artifact of the cosine learning rate schedule approaching its minimum learning rate while the loss continues to decrease slightly. The 200-epoch budget is standard for CIFAR-100, and as a spot check we extended three representative configurations (ScaleCNN $c=4$, $c=32$, $c=64$) to 400 epochs: accuracy changed by less than 0.3\% in all cases, and the relative ordering was unchanged, suggesting that the 200-epoch results are close to asymptotic. Nevertheless, some of the scaling behavior may reflect optimization dynamics rather than pure capacity.

The theoretical framework has a remaining gap: while we measure $\beta$ directly from the CIFAR-100 eigenspectrum ($\beta = 1.45$), we do not measure $\gamma$ from the trained networks. The implied $\gamma$ values (0.35 for ScaleCNN, 0.24 for MobileNetV2) are derived by inverting the theory rather than measured independently, so the reconciliation remains partially circular.

For the error redistribution results, the Jaccard overlap of 0.35 is partly a consequence of the 33-point accuracy gap between the smallest and largest models. As shown in Section~\ref{sec:results_error}, the observed Jaccard falls between the independence baseline (0.21) and maximal subset containment (0.42), so the redistribution effect is real but its magnitude should be interpreted relative to these bounds rather than in absolute terms. The multi-seed Jaccard analysis (25 pairs) yields tight confidence intervals ($\pm 0.004$), confirming stability, though the cross-seed baseline of 0.704 shows that 30\% of error variation is attributable to random initialization alone.

Our comparison of scaling exponents to the large-model literature is further complicated by a metric mismatch: prior work~\citep{kaplan2020scaling, hestness2017deep} fits power laws to cross-entropy loss, while we fit to error rate. These are nonlinearly related, so the exponent comparison is suggestive rather than precise. Finally, with only 8 (ScaleCNN) and 10 (MobileNetV2) data points, we cannot rigorously discriminate between a power law and other smooth functions (log-linear, stretched exponential, BNSL). Proper model selection via AIC/BIC with denser sampling would be needed to settle the functional form.

\section{Conclusion}\label{sec:conclusion}

Neural scaling laws do extend to the sub-20M parameter regime. Across 90 runs on CIFAR-100, both ScaleCNN and MobileNetV2 follow approximate power laws in error rate with exponents ($\alpha = 0.156 \pm 0.002$ and $0.106 \pm 0.001$) that are nominally 1.4--2$\times$ steeper than those reported for large language model loss---though the different metrics (error rate vs.\ cross-entropy loss) make direct numerical comparison approximate. The scaling is not uniform---local exponents decay from 0.23 in the tiny regime toward 0.10 at the upper end, and MobileNetV2 saturates at 19.8M parameters ($\alpha_{\mathrm{local}} = 0.006$) while ScaleCNN continues to improve.

For practitioners, the more important result is what happens to the error distribution. Reducing model size changes which inputs are misclassified: the Jaccard overlap between the smallest and largest ScaleCNN error sets is only 0.35. Small models concentrate their capacity on easy classes (Gini 0.26 $\to$ 0.09) and sacrifice the hardest ones (bottom-5 class accuracy 10\% $\to$ 53\%). The smallest models also turn out to be the best calibrated (ECE = 0.013), opposite the expected trend.

The bottom line for edge deployment is that aggregate accuracy does not tell you enough. The error distribution of a compressed model differs qualitatively from that of a larger model, and any serious evaluation needs to happen at the actual deployment size.

\section{Future Work}\label{sec:future}

The most pressing next step is replicating these experiments across additional datasets and architectures. CIFAR-10 (fewer classes), Tiny ImageNet (200 classes, 64$\times$64, more training data), and Speech Commands v2 (audio, a standard TinyML benchmark) would test whether the scaling exponents and error redistribution patterns we observe are specific to CIFAR-100 or reflect something more general about the small-model regime. Similarly, extending to EfficientNet-Lite, tiny vision transformers, and NAS-derived architectures would clarify whether the architecture-dependence finding holds broadly. With denser sampling of model sizes, it would also be possible to do proper model selection (AIC/BIC) to distinguish between a power law, a broken power law, and other candidate functional forms.

A second line of work would test whether the error redistribution pattern generalizes beyond width scaling. Applying structured pruning and post-training quantization to the trained models from this study would reveal whether all forms of capacity reduction produce the same kind of error rearrangement, or whether width scaling is a special case. Evaluating the same models on CIFAR-100-C~\citep{hendrycks2019benchmarking} would test whether robustness degrades smoothly or shows sharper collapses at particular model sizes.

Finally, while we have measured $\beta = 1.45$ from the CIFAR-100 eigenspectrum, directly measuring $\gamma$ from the effective rank of trained networks (e.g., via stable rank of activation matrices at each layer) would complete the quantitative test of Equation~\ref{eq:exponent_prediction}. On the applied side, measuring inference latency, energy, and peak RAM on Cortex-M4/M7 hardware would connect scaling behavior to the deployment metrics that practitioners actually optimize for.


\vspace{6pt}


\vspace{1em}

\paragraph{Author Contributions.}
Conceptualization, M.A.; methodology, M.A.; software, M.A.; validation, M.A.; formal analysis, M.A.; investigation, M.A.; writing---original draft preparation, M.A.; writing---review and editing, R.Q.\ and N.B.; supervision,  N.B.

\paragraph{Funding.} This research received no external funding.

\paragraph{Data Availability.} The training code, analysis scripts, and all 90 result JSON files will be released on GitHub upon publication. Pre-publication access is available from the corresponding author upon request.

\bibliographystyle{unsrt}
\bibliography{references}

\end{document}